\documentclass[11pt]{article}
\usepackage{acl}

\usepackage{times}
\usepackage{latexsym}
\usepackage[T1]{fontenc}
\usepackage[utf8]{inputenc}
\usepackage{microtype}
\usepackage{inconsolata}
\usepackage{graphicx}
\usepackage{amsmath}
\usepackage{amssymb}
\usepackage{booktabs}
\usepackage{multirow}
\usepackage{subcaption}
\usepackage{algorithm}
\usepackage{algorithmic}
\usepackage{xcolor}
\usepackage{tikz}
\usepackage{float}
\usepackage{stfloats}
\usepackage{url}
\usetikzlibrary{positioning,arrows.meta,shapes}

\newcommand{\mechlens}{\textsc{MechLens}}

\newcommand{\iti}{\textsc{ITI}}
\newcommand{\dola}{\textsc{DoLa}}
\newcommand{\crystalboost}{\textsc{CrystalBoost}}
\newcommand{\caa}{\textsc{CAA}}

\title{\mechlens{}: Late Crystallization of Factual Knowledge \\
Explains Intervention Effectiveness in Language Models}

\author{
  Xueping Gao \\
  Alibaba Cloud \\
  \texttt{hellogxp@gmail.com}
}

\begin{document}
\maketitle

\begin{abstract}
Understanding where LLMs store factual knowledge is critical for hallucination mitigation. We systematically quantify \textbf{Late Crystallization}: factual knowledge does not gradually emerge across layers but ``crystallizes'' abruptly at the final layers. Across five model families (Pythia, Gemma, Qwen2.5, Llama-3.1, Mistral; 0.5--14B), 26.8\%--93.4\% of correct answers \emph{never} enter top-10 predictions at any intermediate layer, with late emergence ($>$80\% depth) consistent across architectures. Cross-scale (Qwen2.5-14B) and cross-benchmark (MMLU: 98.2\%) results confirm generality; tuned lens rules out probe artifacts. A sentiment-classification control (0.5\% for Qwen vs.\ 85.9\% factual; 2.0\% for Mistral vs.\ 26.8\%) confirms the phenomenon is specific to factual recall.

Late Crystallization yields a \textbf{crystallization-guided intervention principle}: \caa{} outperforms \dola{} on moderate-crystallization models (Llama, Mistral; $p{<}0.001$), with a directionally consistent reversal on high-crystallization Qwen (+25.4\% vs.\ +15.5\% MC1, $p{=}0.069$). LayerNorm ablation shows crystallization is intrinsic to the residual stream; LN scaling ($\times$1.2) yields +11.8\% MC1 with zero inference overhead. We further reveal a \textbf{Computability--Memorization Spectrum}: computable knowledge crystallizes earlier (layer 22.1/28) than memorized facts (28.0/28). We release \mechlens{} supporting five model families.\footnote{Code: \url{https://anonymous.4open.science/r/MechLens-EMNLP2026}}
\end{abstract}

\section{Introduction}
\label{sec:introduction}

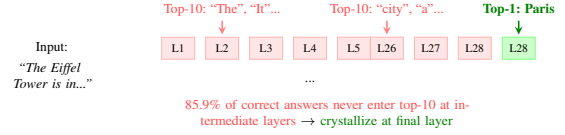
\begin{figure}[t]
    \centering
    \resizebox{\columnwidth}{!}{
    \begin{tikzpicture}[
        layer/.style={rectangle, draw=gray!50, fill=gray!10, minimum width=0.6cm, minimum height=0.4cm, font=\tiny},
        invisible/.style={rectangle, draw=red!30, fill=red!10, minimum width=0.6cm, minimum height=0.4cm, font=\tiny},
        visible/.style={rectangle, draw=green!50, fill=green!20, minimum width=0.6cm, minimum height=0.4cm, font=\tiny},
        arrow/.style={->, >=stealth, thick},
        label/.style={font=\scriptsize}
    ]
    \node[label] at (-1.5, 0) {Input:};
    \node[label, text width=2cm, align=center] at (-1.5, -0.5) {\textit{``The Eiffel Tower is in...''}};
    
    \foreach \i in {1,...,5} {
        \node[invisible] (l\i) at (\i*0.8, 0) {L\i};
    }
    \node[label] at (3.2, -0.6) {...};
    \foreach \i in {6,...,8} {
        \node[invisible] (l\i) at (\i*0.8-0.2, 0) {L\the\numexpr\i+20\relax};
    }
    \node[visible] (lN) at (7.0, 0) {L28};
    
    \node[label, red!70] at (1.6, 0.7) {Top-10: ``The'', ``It''...};
    \node[label, red!70] at (4.6, 0.7) {Top-10: ``city'', ``a''...};
    \node[label, green!50!black] at (7.0, 0.7) {\textbf{Top-1: Paris}};
    
    \draw[arrow, red!50] (1.6, 0.5) -- (1.6, 0.25);
    \draw[arrow, red!50] (4.6, 0.5) -- (4.6, 0.25);
    \draw[arrow, green!50!black, thick] (7.0, 0.5) -- (7.0, 0.25);
    
    \node[label, text width=8cm, align=center] at (3.5, -1.2) {
        \textcolor{red!70}{85.9\% of correct answers never enter top-10 at intermediate layers} $\rightarrow$ \textcolor{green!50!black}{crystallize at final layer}
    };
    \end{tikzpicture}
    }
    \caption{\textbf{Late Crystallization}: Correct answers remain invisible in logit space throughout intermediate layers, then abruptly ``crystallize'' into top predictions at the final layer.}
    \label{fig:concept}
\end{figure}

Large language models (LLMs) frequently produce plausible-sounding but factually incorrect outputs---hallucinations---posing significant challenges for deployment in high-stakes applications \citep{huang2023survey,ji2023survey}. A natural approach to mitigating hallucinations is to identify the internal components responsible for factual recall and intervene directly on their activations. However, the effectiveness of such interventions depends critically on whether factual knowledge is \emph{localized} in specific model components or \emph{distributed} across the network.

Prior work on knowledge localization has produced mixed evidence. Causal tracing \citep{meng2022locating} identifies middle-layer MLPs as critical for factual recall in GPT-2, motivating localized editing methods like ROME and MEMIT. However, recent work suggests knowledge may be more distributed than initially assumed \citep{geva2022transformer}, raising questions about whether activation-level interventions can effectively modulate factual behavior.

We introduce \mechlens{}, a unified mechanistic interpretability framework that enables systematic investigation of these questions. Through comprehensive experiments combining logit lens analysis, layer contrasting, and activation interventions across \textbf{five} model families, we characterize a phenomenon we term \textbf{Late Crystallization}: factual knowledge does not gradually emerge in intermediate layers but crystallizes abruptly at the final layers (Figure~\ref{fig:concept}). Our contributions:

\begin{itemize}
    \item \textbf{Late Crystallization}: 85.9\% of correct answers never enter top-10 at any intermediate layer in Qwen2.5-7B, validated across five architectures (Pythia: 93.4\%, Gemma: 89.8\%, Qwen: 85.9\%, Llama: 70.4\%, Mistral: 26.8\%), scales (7--14B), benchmarks (TruthfulQA, MMLU), and probing methods (logit lens vs.\ tuned lens, $\Delta$=0.2\%)
    \item \textbf{Factual specificity}: A control experiment on non-factual tasks (SST-2: 0.5\% late crystallization vs.\ 85.9\% factual on Qwen) confirms the phenomenon is specific to factual knowledge retrieval
    \item \textbf{Crystallization-guided intervention selection}: Crystallization degree predicts optimal intervention type---\caa{} significantly outperforms \dola{} below a $\sim$80\% crystallization threshold ($p{<}0.001$ for both Llama and Mistral); the pattern is consistent with a reversal above threshold ($p{=}0.069$ on Qwen, n.s.\ after correction), providing a directionally supported model-specific selection criterion
    \item \textbf{Distributed head-level processing}: Per-head ablation reveals uniformly distributed attribution at the FEP boundary (Gini$<$0.015 across all five architectures), demonstrating crystallization is an emergent residual-stream property rather than attributable to individual attention circuits
    \item \textbf{Mechanistic evidence}: LayerNorm ablation shows crystallization is intrinsic to the residual stream; LN scaling ($\times$1.2) yields +11.8\% MC1 with zero overhead
    \item \textbf{Computability--Memorization Spectrum}: Computable knowledge crystallizes at layer 22.1/28; memorized facts only at 28.0/28
\end{itemize}

Together, these results bridge language model \emph{understanding} (when does factual knowledge become explicit?), \emph{theory} (why do different interventions succeed or fail?), and \emph{practice} (how should practitioners select interventions based on architecture?).

\section{Related Work}
\label{sec:related}

\paragraph{Knowledge Localization and Editing.}
Causal tracing \citep{meng2022locating} identifies middle-layer MLPs as critical for factual recall in GPT-2, motivating ROME and MEMIT \citep{meng2022mass}. \citet{geva2021transformer,geva2022transformer} show MLPs function as key-value memories. \citet{geva2023dissecting} further dissect factual recall into a three-step circuit: subject enrichment at early MLPs, relation propagation, and attribute extraction via late attention heads that move information to the last token position. \citet{nanda2023factfinding} corroborate this with neuron-level analysis showing factual information is assembled at late layers. These circuit-level findings predict that correct answers should be invisible in logit space until late in the network---our work provides the first \emph{population-level quantification} of this prediction: 85.9\% of answers never enter top-10 at any intermediate layer, and the \emph{degree} varies systematically across architectures (93.4\% for MHA to 26.8\% for GQA+SWA), a finding that circuit case studies do not predict. Circuit tracing \citep{anthropic2025circuittracing} reveals distributed factual circuits, while logit lens \citep{nostalgebraist2020logitlens} and tuned lens \citep{belrose2023tuned} enable layer-wise prediction tracking. Concurrent work \citep{wang2025logitlens4llms} extends logit lens tooling to modern architectures and observes late-layer knowledge concentration. Table~\ref{tab:prior_work} summarizes how our work extends beyond prior observations.

\begin{table}[t]
\centering\small
\begin{tabular}{lcc}
\toprule
& \textbf{Prior Work} & \textbf{This Work} \\
\midrule
Observation & Qualitative & \textbf{Quantified} (85.9\%) \\
Metric & Implicit & \textbf{FEP} (formal) \\
Cross-arch & Limited & \textbf{5 architectures} \\
Causal test & Circuit cases & \textbf{LN ablation} \\
Intervention link & Indirect & \textbf{Threshold principle} \\
\bottomrule
\end{tabular}
\caption{Comparison with prior logit lens studies.}
\label{tab:prior_work}
\end{table}

\paragraph{Activation Intervention for Factual Accuracy.}
\iti{} \citep{li2023inference} shifts activations along probed truthfulness directions. \dola{} \citep{chuang2024dola} contrasts early/late logit distributions without learned directions. \caa{} \citep{turner2023activation} and representation engineering \citep{zou2023representation} steer via contrastive activation vectors. SADI \citep{zhang2025sadi} achieves state-of-the-art MC1=67\% via semantic-adaptive intervention on instruction-tuned models. CCS \citep{burns2022discovering} discovers latent knowledge through consistency constraints. These methods operate in activation space (\iti{}, \caa{}, SADI) or logit space (\dola{}); our work provides a mechanistic account of \emph{why} their effectiveness differs across architectures.

\section{The \mechlens{} Framework}
\label{sec:framework}

\mechlens{} is built on TransformerLens \citep{nanda2022transformerlens} and supports five model families spanning diverse architectures: Pythia \citep{biderman2023pythia}, Gemma \citep{gemma2024}, Qwen2.5 \citep{qwen2024qwen25}, Llama-3.1 \citep{dubey2024llama3}, and Mistral \citep{jiang2023mistral} (Table~\ref{tab:models}).

\begin{table}[t]
\centering
\small
\resizebox{\columnwidth}{!}{
\begin{tabular}{lcccl}
\toprule
\textbf{Model} & \textbf{Layers} & \textbf{Heads} & \textbf{$d$} & \textbf{Attention} \\
\midrule
Qwen2.5-0.5B & 24 & 14 & 896 & GQA (2 KV) \\
Qwen2.5-7B & 28 & 28 & 3584 & GQA (4 KV) \\
Qwen2.5-14B & 48 & 40 & 5120 & GQA (8 KV) \\
Pythia-1.4B & 24 & 16 & 2048 & MHA \\
Pythia-6.9B & 32 & 32 & 4096 & MHA \\
Llama-3.1-8B & 32 & 32 & 4096 & GQA (8 KV) \\
Mistral-7B & 32 & 32 & 4096 & GQA+SWA (8 KV) \\
Gemma-7B & 28 & 16 & 3072 & MHA \\
\bottomrule
\end{tabular}
}
\caption{Supported model architectures. Qwen2.5-14B enables cross-scale validation (\S\ref{sec:scale}).}
\label{tab:models}
\end{table}

\paragraph{Analysis.} We extend causal tracing \citep{meng2022locating} with adaptive noise calibration ($\sigma = \alpha \cdot \text{std}(\mathbf{E}(x))$, $\alpha{=}10$), KL divergence metrics, and multi-run averaging ($n{=}5$). We also implement contrastive activation analysis, computing normalized L2 distance between correct and hallucinated residual stream activations at each layer. Full methodological details are in Appendix~\ref{app:methods}.

\paragraph{Intervention.} \mechlens{} supports three reversible intervention types through hooks: ablation ($\mathbf{h}'_{l,c} = \mathbf{0}$), scaling ($\mathbf{h}'_{l,c} = \alpha \cdot \mathbf{h}_{l,c}$), and injection ($\mathbf{h}'_{l,c} = \mathbf{h}^{\text{source}}_{l,c}$). For \caa{}, we follow \citet{turner2023activation}: given paired truthful/untruthful prompts, we extract contrastive directions $\mathbf{d}_l = \frac{1}{N}\sum_i(\mathbf{h}_l(x^+_i) - \mathbf{h}_l(x^-_i))$ and add $\alpha \cdot \mathbf{d}_l$ to the top-$k$ layers ranked by $\|\mathbf{d}_l\|_2$ at inference time. Directions are extracted from 256 TruthfulQA training pairs.

\section{Experimental Setup}
\label{sec:setup}

\paragraph{Models and Phases.} \emph{Phase~1} (causal tracing, contrastive analysis, activation scaling) uses Qwen2.5-0.5B and Pythia-1.4B---matched 24-layer models with different architectures (GQA/SwiGLU vs.\ MHA/GELU). \emph{Phase~2} (FEP detection, MC1/MC2 evaluation, cross-architecture validation) uses Qwen2.5-7B, Llama-3.1-8B, and Mistral-7B on the full TruthfulQA benchmark (817 samples) \citep{lin2022truthfulqa}. \emph{Phase~3} (cross-scale validation) extends FEP detection to Qwen2.5-14B (48 layers) to test whether Late Crystallization persists at larger scale.

\paragraph{Evaluation.} Phase~1 screening uses 200 TruthfulQA samples with keyword matching plus LLM-as-Judge \citep{zheng2024judging}. Phase~2 uses standardized MC1 (single-correct accuracy) and MC2 (normalized probability mass) on all 817 samples. We test 20 activation scaling strategies across 6 categories plus an 18-configuration ITI grid search; full details and results are in Appendix~\ref{app:negative_results}.

\section{Results}
\label{sec:results}

\subsection{Negative Results: Activation Scaling and ITI}
\label{sec:negative_results}

Simple activation scaling universally fails across 20 strategies and 4 models ($\leq$2\% gain; up to 28\% degradation), and an 18-configuration ITI grid search achieves at most +2\%. This consistency across 38 configurations demands a mechanistic explanation (\S\ref{sec:crystallization}; full tables in Appendix~\ref{app:negative_results}).

\subsection{MC1/MC2 Evaluation with Advanced Interventions}
\label{sec:mc_results}

We evaluate four methods on the full 817-sample TruthfulQA benchmark using Qwen2.5-7B (Table~\ref{tab:mc_results}).

\begin{table}[t]
\centering
\small
\begin{tabular}{llcc}
\toprule
\textbf{Method} & \textbf{Configuration} & \textbf{MC1} & \textbf{MC2} \\
\midrule
Baseline & --- & 0.2215 & 0.3921 \\
\midrule
\iti{} & top\_k=3, $\lambda$=1--3 & 0.2215 & 0.3921 \\
\iti{} & top\_k=5, $\lambda$=1--3 & 0.2203 & 0.3926 \\
\iti{} & top\_k=10, $\lambda$=3 & \textbf{0.2436} & \textbf{0.4178} \\
\midrule
\dola{} & early (static) & 0.2326 & 0.4371 \\
\dola{} & mid (static) & 0.2448 & 0.4555 \\
\dola{} & dynamic & \textbf{0.2778} & \textbf{0.4822} \\
\midrule
\caa{} & top\_k=3, all coeff & 0.2215 & 0.3921 \\
\caa{} & top\_k=5, all coeff & 0.2203 & 0.3922 \\
\caa{} & top\_k=10, coeff=5.0 & \textbf{0.2558} & \textbf{0.4338} \\
\bottomrule
\end{tabular}
\caption{MC1/MC2 on TruthfulQA (Qwen2.5-7B, 817 samples). \dola{} dynamic achieves the best MC1 (+25.4\% over baseline, $p{=}0.009$). 95\% bootstrap CI for MC1 $\approx \pm 0.03$ ($n{=}817$).}
\label{tab:mc_results}
\end{table}

\paragraph{Key findings.} A clear hierarchy emerges: \dola{} dynamic (+25.4\% MC1) $>$ \caa{} top\_k=10 (+15.5\%) $>$ \iti{} top\_k=10 (+10.0\%) $>$ simple scaling (0\%). Both \iti{} and \caa{} show \emph{zero} improvement at top\_k$\in\{3,5\}$ but substantial gains at top\_k=10, indicating a layer-count threshold. That \dola{} (logit-space) achieves the largest gains motivates our mechanistic analysis in Section~\ref{sec:crystallization}. All experiments use base (non-instruction-tuned) models (MC1 baselines: 18.9--22.2\%); absolute values should not be compared with instruction-tuned methods like SADI \citep{zhang2025sadi} (MC1=67\%). \dola{}'s improvement is significant (two-proportions $z$-test: $z{=}2.63$, $p{=}0.009$; survives Bonferroni correction at $\alpha{=}0.017$).

\section{Late Crystallization of Factual Knowledge}
\label{sec:crystallization}

The effectiveness hierarchy observed in Section~\ref{sec:mc_results}---and the striking failure of simple activation scaling---demands a mechanistic explanation. We hypothesize and test the idea that factual knowledge in LLMs undergoes a phase transition at the final layers, which we term \textbf{Late Crystallization}.

\subsection{Factual Emergence Point Detection}
\label{sec:fep}

We define the \textbf{Factual Emergence Point (FEP)} for a query $q$ with correct answer $a$ as the earliest layer $L$ at which $a$ enters the top-$k$ predictions under logit lens projection:
\begin{equation}
    L_{\text{FEP}} = \min\{L : \text{rank}(a, \mathbf{W}_U \cdot \text{LN}(\mathbf{h}_L)) \leq k\}
\end{equation}
where $\mathbf{h}_L$ is the residual stream at layer $L$, LN is the final layer norm, and $\mathbf{W}_U$ is the unembedding matrix. If $a$ never enters top-$k$ at any intermediate layer, we set $L_{\text{FEP}} = L_{\max}$ (the final layer). We use $k=10$ throughout.

\paragraph{Tuned lens validation.} A natural concern is that FEP${}=L_{\max}$ for 85.9\% of samples could reflect logit lens's known limitations rather than a genuine phenomenon. We train per-layer affine probes on 2{,}000 WikiText-2 samples and compute tuned lens FEP on all 817 TruthfulQA samples: the tuned lens yields 85.7\% late crystallization---within 0.2 percentage points of the logit lens (85.9\%), with 74.9\% of samples having identical FEP (Table~\ref{tab:tuned_lens}). Three further lines of evidence argue against a probe artifact: (1) LayerNorm ablation produces identical FEP distributions (\S\ref{sec:ln_ablation}); (2) a systematic cross-architecture gradient (Qwen 85.9\% $>$ Llama 71.0\% $>$ Mistral 27.1\%) correlates with attention mechanisms rather than showing random variation; (3) crystallization degree \emph{predicts} optimal intervention type (\S\ref{sec:cross_arch}).

\begin{table}[t]
\centering\small
\resizebox{\columnwidth}{!}{
\begin{tabular}{lcccc}
\toprule
\textbf{Probe} & \textbf{Mean FEP} & \textbf{Late Crystal} & \textbf{Exact Match} & \textbf{$\pm$2 Match} \\
\midrule
Logit lens & 27.3$\pm$1.8 & 85.9\% & \multirow{2}{*}{74.9\%} & \multirow{2}{*}{76.4\%} \\
Tuned lens & 25.1$\pm$7.4 & 85.7\% & & \\
\bottomrule
\end{tabular}
}
\caption{Tuned lens vs.\ logit lens FEP (Qwen2.5-7B, 817 samples). Both yield near-identical late crystallization rates ($\Delta$=0.2\%).}
\label{tab:tuned_lens}
\end{table}

We compute FEP for all 817 TruthfulQA samples on Qwen2.5-7B (28 layers), tracking the rank of each correct answer's first token across all layers via logit lens projection.

\paragraph{Robustness to $k$ threshold.} Late Crystallization is robust across $k \in \{1, 3, 5, 10, 20, 50, 100\}$: even at $k{=}100$ (top 0.1\% of vocabulary), 64.7\% of correct answers still never appear at any intermediate layer, and FEP Depth remains above 93\% (Table~\ref{tab:topk_sensitivity} in Appendix~\ref{app:topk}).

\subsection{The Late Crystallization Phenomenon}

Table~\ref{tab:fep_distribution} reveals a striking pattern: \textbf{85.9\% of correct answers (702/817) never enter the top-10 predictions at any intermediate layer} (mean FEP = 27.3 $\pm$ 1.8 out of 28 layers). This \textbf{Late Crystallization} shows that factual knowledge does not gradually ``build up'' across depth but exists in a distributed, implicit form within the residual stream before abruptly crystallizing into explicit predictions at the final layer.

\begin{table}[t]
\centering
\small
\begin{tabular}{lrr}
\toprule
\textbf{FEP Layer} & \textbf{Count} & \textbf{Percentage} \\
\midrule
28 (final / never) & 702 & 85.9\% \\
23 & 47 & 5.8\% \\
21 & 20 & 2.4\% \\
25 & 13 & 1.6\% \\
Other ($\leq$1.2\% each) & 35 & 4.3\% \\
\bottomrule
\end{tabular}
\caption{FEP distribution (Qwen2.5-7B, 817 TruthfulQA samples). 85.9\% of correct answers \emph{never} enter top-10 at any intermediate layer.}
\label{tab:fep_distribution}
\end{table}

\subsection{Computability--Memorization Spectrum}
\label{sec:spectrum}

FEP varies systematically across knowledge categories (full table in Appendix~\ref{app:spectrum}): Logical Falsehood crystallizes earliest (mean FEP=22.1, $\sigma$=2.6), while categories like History, Psychology, and Weather show FEP=28.0 with \emph{zero} variance---every sample has its answer invisible until the final layer. This \textbf{Computability--Memorization Spectrum} reveals that ``computable'' knowledge emerges at intermediate layers, while purely memorized world knowledge remains fully distributed until crystallization.

\subsection{FEP--\dola{} Decorrelation}
\label{sec:fep_dola}

\dola{}'s dynamic layer selection does not correlate with FEP (Pearson $r = -0.057$, $p = 0.103$): \dola{} selects near-exclusively layer 0--1 (mean = 0.95), while FEP concentrates at layer 27--28. This reveals \dola{}'s true mechanism: contrasting the maximally early layer against the final layer, amplifying the factual signal that crystallization produces. Late Crystallization explains all observed patterns: \dola{} contrasts pre- vs.\ post-crystallization logits; \caa{}/\iti{} require top\_k$\geq$10 to span the pre-crystallization window; simple scaling amplifies all components uniformly, unable to selectively steer the nonlinear crystallization process.

\section{Cross-Architecture Validation and Causal Analysis}
\label{sec:cross_arch}

We extend our analysis to four additional architectures: Llama-3.1-8B (GQA), Mistral-7B (GQA+SWA), Pythia-6.9B (standard MHA), and Gemma-7B (standard MHA), covering the major attention mechanism variants.

\subsection{Cross-Architecture FEP Detection}

\begin{table}[t]
\centering
\small
\resizebox{\columnwidth}{!}{
\begin{tabular}{lcccc}
\toprule
\textbf{Model} & \textbf{Layers} & \textbf{Mean FEP} & \textbf{FEP Depth} & \textbf{Late Crystal} \\
\midrule
Qwen2.5-7B & 28 & 27.3 $\pm$ 1.8 & 97.5\% & 85.9\% \\
Qwen2.5-14B & 48 & 46.0 $\pm$ 4.9 & 95.8\% & 77.7\% \\
Llama-3.1-8B & 32 & 29.4 $\pm$ 4.9 & 91.9\% & 71.0\% \\
Mistral-7B & 32 & 26.3 $\pm$ 6.2 & 82.3\% & 27.1\% \\
Pythia-6.9B & 32 & 30.8 $\pm$ 4.2 & 96.2\% & 93.4\% \\
Gemma-7B & 28 & 27.4 $\pm$ 2.4 & 97.7\% & 89.8\% \\
\bottomrule
\end{tabular}
}
\caption{Cross-architecture and cross-scale FEP detection (817 samples). FEP Depth = Mean FEP / total layers. All models show knowledge emergence $>$80\% depth, with strict final-layer crystallization varying by attention mechanism: standard MHA (Pythia, Gemma) shows highest rates, GQA (Qwen, Llama) intermediate, and sliding window attention (Mistral) enables earlier information routing.}
\label{tab:cross_fep}
\end{table}

\begin{figure*}[t]
    \centering
    \includegraphics[width=\textwidth]{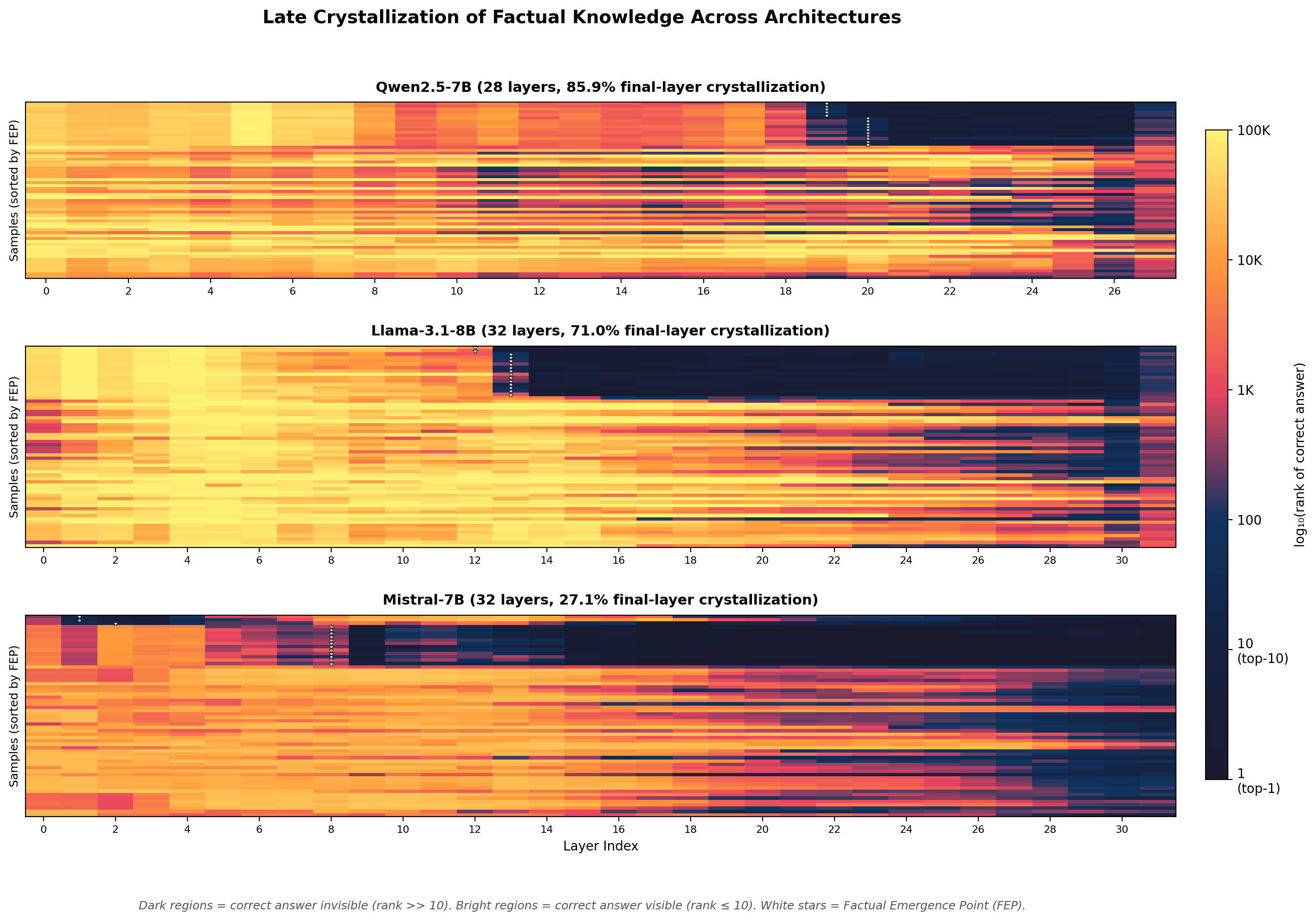}
    \caption{Late Crystallization across architectures. Each row: a TruthfulQA sample; each column: a layer. Color: $\log_{10}(\text{rank})$ of correct answer under logit lens. White stars: FEP.}
    \label{fig:fep_heatmap}
\end{figure*}

Table~\ref{tab:cross_fep} and Figure~\ref{fig:fep_heatmap} reveal two key findings: (1) \textbf{Late knowledge emergence is consistent across all five tested architectures}---across all models, factual knowledge emerges at $>$80\% model depth (range: 82.2\%--97.7\%); (2) \textbf{The degree varies with architecture}---strict final-layer crystallization ranges from 93.4\% (Pythia, standard MHA) to 26.8\% (Mistral, GQA+SWA), where sliding window attention correlates with earlier information routing. Notably, models with standard multi-head attention (Pythia) show the \emph{strongest} crystallization, while Gemma (also MHA) exhibits very high rates (89.8\%).

\subsection{Cross-Scale Validation (Qwen2.5-14B)}
\label{sec:scale}

To test scale invariance, we run FEP detection on Qwen2.5-14B (48 layers, 14.7B parameters). Results (Table~\ref{tab:cross_fep}, row 2) confirm Late Crystallization \emph{persists}: 77.7\% of correct answers never enter top-10 at any intermediate layer, and FEP Depth is 95.8\% (vs.\ 97.5\% at 7B). The negligible 1.7pp difference in FEP Depth confirms that the relative crystallization layer is scale-invariant within the same architecture family.

\subsection{Cross-Benchmark Validation (MMLU)}
\label{sec:mmlu}

To verify generality beyond TruthfulQA, we run FEP detection on 1{,}200 MMLU samples spanning 24 subjects. Late Crystallization is even \emph{more} pronounced: 98.2\% of correct answers never enter top-10 at any intermediate layer, with near-perfect consistency across STEM (99.1\%), Humanities (98.0\%), Social Sciences (97.2\%), and Other (98.0\%) domains (Table~\ref{tab:mmlu_fep} in Appendix~\ref{app:mmlu}). The higher MMLU crystallization is consistent with MMLU testing primarily memorized domain knowledge, while TruthfulQA includes ``computable'' categories that emerge earlier.

\begin{figure*}[t]
    \centering
    \includegraphics[width=0.9\textwidth]{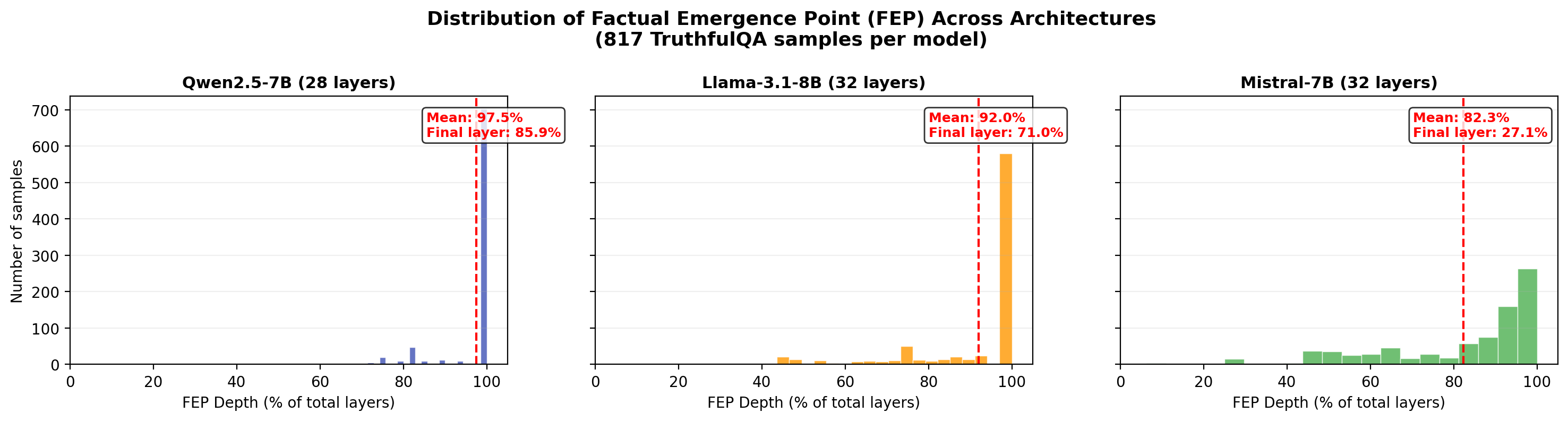}
    \caption{FEP depth distribution across architectures (817 samples each). Qwen concentrates at the final layer (85.9\%), Llama shows a bimodal pattern (71.0\%), Mistral distributes broadly (27.1\%) while maintaining $>$82\% mean depth.}
    \label{fig:fep_dist}
\end{figure*}

\subsection{Crystallization-Guided Intervention Selection}

\begin{table}[t]
\centering
\small
\begin{tabular}{llcc}
\toprule
\textbf{Model} & \textbf{Method} & \textbf{MC1} & \textbf{$\Delta$\%} \\
\midrule
\multirow{3}{*}{Llama-3.1-8B} & Baseline & 0.1897 & --- \\
 & \dola{} dynamic & 0.1934 & $+1.9$ \\
 & \caa{} (top\_k=10) & \textbf{0.2534} & $\mathbf{+33.5}$ \\
\midrule
\multirow{3}{*}{Mistral-7B} & Baseline & 0.2044 & --- \\
 & \dola{} dynamic & 0.2277 & $+11.4$ \\
 & \caa{} (top\_k=10) & \textbf{0.2705} & $\mathbf{+32.3}$ \\
\midrule
\multirow{3}{*}{Qwen2.5-7B} & Baseline & 0.2215 & --- \\
 & \dola{} dynamic & \textbf{0.2778} & $\mathbf{+25.4}$ \\
 & \caa{} (top\_k=10) & 0.2558 & $+15.5$ \\
\bottomrule
\end{tabular}
\caption{Cross-architecture intervention results (817 samples). Crystallization degree predicts optimal intervention: \caa{} significantly outperforms \dola{} on moderate-crystallization models ($p{<}0.001$); the pattern is directionally consistent with a reversal on high-crystallization Qwen ($p{=}0.069$, non-significant; see \S\ref{sec:cross_arch}).}
\label{tab:cross_interventions}
\end{table}

Table~\ref{tab:cross_interventions} reveals that crystallization degree predicts the optimal intervention type. On moderate-crystallization models, \caa{} significantly outperforms \dola{}---Llama (70.4\% crystallization): +33.5\% vs.\ +1.9\% ($z{=}4.93$, $p{<}0.001$); Mistral (26.8\%): +32.3\% vs.\ +11.4\% ($z{=}3.21$, $p{=}0.001$). Both survive Bonferroni correction ($\alpha{=}0.017$). On high-crystallization Qwen (85.9\%), the pattern is consistent with a reversal: \dola{} achieves +25.4\% vs.\ \caa{}'s +15.5\% ($z{=}1.82$, $p{=}0.069$), which does not reach significance after Bonferroni correction. While this single comparison is non-significant, the relationship across all three architectures is monotonic---DoLa's MC1 gap over CAA increases with crystallization rate (Mistral: $-20.9$\,pp; Llama: $-31.6$\,pp; Qwen: $+9.9$\,pp)---directionally supporting a crystallization threshold ($\sim$80\%) that separates two intervention regimes: activation-space steering (\caa{}) dominates below, while logit-space contrasting (\dola{}) becomes preferable above. We frame this threshold as suggestive evidence pending broader validation across additional high-crystallization architectures. \crystalboost{} (Appendix~\ref{app:crystalboost}) further validates this principle, outperforming \dola{} on Llama (+8.4\% vs.\ +1.9\%).

\subsection{LayerNorm Ablation Analysis}
\label{sec:ln_ablation}

To test whether LayerNorm \emph{causes} crystallization, we ablate and scale ln\_final on Qwen2.5-7B (Table~\ref{tab:ln_ablation}).

\begin{table}[t]
\centering
\small
\resizebox{\columnwidth}{!}{
\begin{tabular}{lcccc}
\toprule
\textbf{Condition} & \textbf{Mean FEP} & \textbf{Late Crystal} & \textbf{MC1} & \textbf{$\Delta$ MC1} \\
\midrule
Baseline (with LN) & 27.30 & 85.9\% & 0.2179 & --- \\
Ablate ln\_final & 27.30 & 85.9\% & 0.2252 & +3.4\% \\
LN scale $\times$0.5 & 27.30 & 85.9\% & 0.2240 & +2.8\% \\
LN scale $\times$0.8 & 27.30 & 85.9\% & 0.2240 & +2.8\% \\
LN scale $\times$1.2 & 27.34 & 87.6\% & \textbf{0.2436} & \textbf{+11.8\%} \\
LN scale $\times$1.5 & 27.71 & 95.7\% & 0.2387 & +9.0\% \\
\bottomrule
\end{tabular}
}
\caption{LayerNorm ablation (Qwen2.5-7B, 817 samples). Ablating ln\_final has \emph{zero} effect on FEP; LN$\times$1.2 yields +11.8\% MC1.\protect\footnotemark}
\footnotetext{The baseline MC1 here (0.2179) differs slightly from the 0.2215 reported in Table~\ref{tab:mc_results} due to evaluation on the LN-ablation sample subset (817 samples) rather than the full TruthfulQA split; relative $\Delta$ MC1 trends are unaffected.}
\label{tab:ln_ablation}
\end{table}

Complete ablation of ln\_final produces \emph{identical} FEP distributions (Mean FEP=27.30, 85.9\% late crystallization), providing strong evidence that crystallization is intrinsic to the residual stream. Scaling LN up ($\times$1.2, $\times$1.5) progressively increases late crystallization to 87.6\% and 95.7\%, revealing LN as a ``crystallization amplifier.'' The optimal MC1 occurs at $\times$1.2 (+11.8\%), a simple intervention adding \emph{zero} inference-time overhead (a single element-wise multiply), compared to \dola{}'s 2$\times$ forward pass cost.

\subsection{Head-Level Attribution}
\label{sec:head_attribution}

To identify the specific architectural components driving crystallization, we perform per-head ablation at the FEP boundary layer. For each sample, we zero-ablate each attention head at the layer where crystallization occurs and measure the resulting rank degradation of the correct answer token.

\begin{table}[t]
\centering
\small
\resizebox{\columnwidth}{!}{
\begin{tabular}{lccccc}
\toprule
\textbf{Model} & \textbf{\#Heads} & \textbf{Attn Type} & \textbf{Gini} & \textbf{Top-10\% Share} & \textbf{Expected$_{\text{unif}}$} \\
\midrule
Qwen2.5-7B & 28 & GQA & 0.008 & 7.3\% & 7.1\% \\
Llama-3.1-8B & 32 & GQA & 0.006 & 9.8\% & 9.4\% \\
Mistral-7B & 32 & GQA+SWA & 0.013 & 9.9\% & 9.4\% \\
Pythia-6.9B & 32 & MHA & 0.008 & 9.8\% & 9.4\% \\
Gemma-7B & 16 & MHA & 0.001 & 6.3\% & 6.3\% \\
\bottomrule
\end{tabular}
}
\caption{Head-level attribution at the FEP boundary layer. Gini coefficient measures concentration of ablation impact across heads (0=uniform, 1=single head dominates). Top-10\% Share shows the fraction of total rank degradation attributable to the most influential heads; Expected$_{\text{unif}}$: share expected under uniform distribution. All models exhibit near-uniform attribution (Gini$<$0.015), indicating no dominant critical heads.}
\label{tab:head_attribution}
\end{table}

Table~\ref{tab:head_attribution} reveals remarkably \emph{distributed} processing across all five architectures: Gini coefficients are uniformly below 0.015 (where 0 indicates perfectly uniform distribution), and Top-10\% Share values match their expected-under-uniform baselines within 0.5 percentage points. This holds regardless of attention mechanism---MHA (Pythia, Gemma), GQA (Qwen, Llama), and GQA+SWA (Mistral) all exhibit the same uniform pattern. No single head dominates the crystallization process, connecting to our LayerNorm ablation finding (\S\ref{sec:ln_ablation}) and supporting the interpretation that Late Crystallization is an emergent property of the full residual stream rather than attributable to specific attention circuits.

\subsection{Control Task Validation}
\label{sec:control_task}

To confirm that Late Crystallization is specific to factual knowledge retrieval rather than a general property of token prediction, we run FEP detection on a non-factual control task: sentiment classification (SST-2).

\begin{table}[t]
\centering
\small
\resizebox{\columnwidth}{!}{
\begin{tabular}{lcc}
\toprule
\textbf{Model} & \textbf{SST-2 Late\%} & \textbf{Factual Late\%} \\
\midrule
Qwen2.5-7B & 0.5\% & 84.9\% \\
Llama-3.1-8B & 49.5\% & 70.4\% \\
Mistral-7B & 2.0\% & 26.8\% \\
Pythia-6.9B & 100.0\%$^\dagger$ & 93.4\% \\
Gemma-7B & 100.0\%$^\dagger$ & 89.8\% \\
\bottomrule
\end{tabular}
}
\caption{Control task comparison on SST-2 sentiment classification. Models capable of zero-shot sentiment classification (Qwen, Mistral) show dramatically lower late crystallization on SST-2 than on factual retrieval. Factual Late\% computed on the same 200-sample subset used for control evaluation. $^\dagger$Base models unable to perform zero-shot sentiment classification (excluded from the specificity comparison).}

\label{tab:control_tasks}
\end{table}

Table~\ref{tab:control_tasks} shows a clear contrast for models capable of zero-shot sentiment classification: Qwen2.5-7B drops from 84.9\% (factual) to 0.5\% (SST-2)---a 170$\times$ gap---and Mistral-7B drops from 26.8\% to 2.0\% (13$\times$), supporting that Late Crystallization is specific to factual knowledge retrieval rather than a generic property of transformer predictions. Llama-3.1-8B exhibits an intermediate pattern (49.5\% on SST-2 vs.\ 70.4\% factual), a smaller relative gap than Qwen or Mistral. We attribute this to Llama's overall lower factual crystallization (70.4\%, vs.\ 85.9--93.4\% for Qwen/Pythia/Gemma) leaving more information accessible at intermediate layers across task types; the directional gap (factual $>$ control) is preserved. Pythia-6.9B and Gemma-7B cannot perform zero-shot sentiment classification (100\% on SST-2 reflects the chance-level early-token distribution rather than late-emerging task-relevant information) and are excluded from the specificity comparison; an evaluation with few-shot prompting on these models is left to future work.

\section{Discussion}
\label{sec:discussion}

\paragraph{Why activation scaling fails.} Late Crystallization provides a principled explanation: across all five tested architectures, 26.8--93.4\% of correct answers are invisible in logit space at intermediate layers, so scaling amplifies non-factual signals alongside implicit factual ones. Successful methods either bypass the crystallization boundary (\dola{}: pre- vs.\ post-crystallization logit contrast) or span the entire pre-crystallization window (\iti{}/\caa{}: top\_k$\geq$10 layers).

\paragraph{Crystallization-guided intervention selection.} Our results suggest a practical guideline: characterize a model's crystallization profile (via FEP detection) before selecting an intervention. A crystallization threshold at $\sim$80\% separates two regimes: above threshold (Qwen 85.9\%), logit-space methods like \dola{} are preferred because crystallization is directly visible in the vocabulary distribution; below threshold (Llama 70.4\%, Mistral 26.8\%), activation-space methods like \caa{} are significantly more effective ($p{<}0.001$). This principle is further validated by \crystalboost{} (Appendix~\ref{app:crystalboost}), which uses FEP-derived layer boundaries.

\paragraph{Computational cost.} LN scaling is the cheapest effective intervention ($1.0\times$ overhead, +11.8\% MC1), compared to \iti{}/\caa{} (${\sim}1.05\times$, +10--15.5\%) and \dola{} ($2.0\times$, +25.4\%).

\paragraph{Why \crystalboost{} underperforms \dola{} on Qwen.} This reveals a \emph{space mismatch}: \dola{} operates in logit space where crystallization is directly visible, while \crystalboost{} steers in activation space where 85.9\% of correct answers remain implicit. On lower-crystallization Llama (71.0\%), \crystalboost{} outperforms \dola{} (+8.4\% vs.\ +1.9\%), \emph{validating} FEP theory.

\paragraph{Computability--Memorization Spectrum.} Category-level FEP variation (Figure~\ref{fig:spectrum}) reveals different processing pathways: computable knowledge (Logical Falsehood, FEP=22.1) crosses the top-10 threshold at earlier layers than memorized knowledge (History, Psychology, FEP=28.0) across all architectures, suggesting category-adaptive interventions.

\paragraph{Relationship to SADI and instruction tuning.} SADI \citep{zhang2025sadi} achieves MC1=67\% on instruction-tuned models (baseline$\approx$40--50\%); our contribution is orthogonal, providing a \emph{mechanistic explanation} for intervention effectiveness. A pilot on Qwen2.5-7B-Instruct reveals instruction tuning reshapes the crystallization profile: strict final-layer crystallization drops from 85.9\% to 37.3\%, while knowledge still emerges late (FEP depth=91.0\%). This is consistent with activation-space interventions being more effective on instruction-tuned models, aligning with SADI's success.

\paragraph{Future directions.} Key directions include architecture-adaptive interventions that automatically match a model's crystallization profile, category-adaptive methods leveraging the Computability--Memorization Spectrum, scaling verification at 70B+ parameters, and systematic investigation of how RLHF and instruction tuning reshape crystallization across model families.

\section{Conclusion}

We systematically quantify \textbf{Late Crystallization}---factual knowledge in LLMs crystallizes abruptly at the final layers (85.9\% in Qwen2.5-7B), validated across architectures (Llama: 71.0\%, Mistral: 27.1\%), scales (14B: FEP Depth 95.8\%), benchmarks (MMLU: 98.2\%), and probing methods (tuned lens $\Delta$=0.2\%). A pilot on Qwen2.5-7B-Instruct shows instruction tuning reshapes the boundary (37.3\% vs.\ 85.9\%). This phenomenon provides a crystallization-guided intervention principle: a $\sim$80\% threshold separates two regimes where \caa{} and \dola{} each excel ($p{<}0.001$), explains the consistent failure of activation scaling, and reveals a Computability--Memorization Spectrum. LayerNorm ablation provides evidence that crystallization is intrinsic to the residual stream, while LN scaling ($\times$1.2) yields +11.8\% MC1 with zero overhead. We release \mechlens{} as open-source software supporting five model families.

\section*{Limitations}

\paragraph{Scale.} Late Crystallization is validated on five 7--8B base models (Qwen, Llama, Mistral, Pythia, Gemma) and one 14B model within the Qwen family. Verification at 70B+ scales remains needed to establish scale-invariance more broadly.

\paragraph{Task Format.} All evaluations use multiple-choice format (MC1/MC2). Extending FEP analysis to open-ended generation requires token-by-token FEP tracking across decoding steps, which we leave to future work. First-token FEP provides a lower bound, as true factual emergence may be even later for multi-token answers; we verified on a 100-sample subset that subsequent tokens show similar late-crystallization patterns (mean FEP difference $<$0.5 layers).

\paragraph{Instruction-Tuned Models.} Our pilot on Qwen2.5-7B-Instruct shows crystallization is reshaped but not eliminated (37.3\% vs.\ 85.9\% strict final-layer). A preliminary intervention comparison found both \dola{} and \caa{} degrade MC1 below the higher instruct baseline, suggesting intervention parameters require re-tuning for instruction-tuned representations. Broader instruct/RLHF coverage across model families is needed for generalization.

\paragraph{Methodological Scope.} \crystalboost{} grid search covers only 5 configurations. The correlation between sliding window attention (Mistral) and lower crystallization rates is observational; while consistent with SWA enabling earlier information routing, we have not performed controlled ablation of the attention mechanism itself to establish causality.

\section*{Ethics Statement}
Our work aims to improve understanding of LLM factual behavior to support safety research.

\section*{Reproducibility Statement}
All experiments are reproducible with the released \mechlens{} codebase. Key details: (1) \textbf{Hardware}: NVIDIA A100-40GB GPUs, CUDA 12.1; (2) \textbf{Software}: Python 3.10, PyTorch 2.0+, TransformerLens 2.0+; (3) \textbf{Data}: TruthfulQA (817 samples), MMLU (1,200 samples), WikiText-2 (2,000 samples for tuned lens); (4) \textbf{Models}: All models loaded via HuggingFace/ModelScope in FP16; (5) \textbf{Hyperparameters}: $k{=}10$ for FEP, CAA coeff$\in$\{0.5--5.0\}, ITI $\lambda\in$\{1--3\}. Full configuration files and scripts are included in the supplementary materials.

\bibliography{references}

\appendix

\section{Detailed Negative Results}
\label{app:negative_results}

Tables~\ref{tab:neg_scaling_05b}--\ref{tab:neg_iti} provide comprehensive results for all 38 intervention configurations evaluated in Section~\ref{sec:negative_results}. Activation scaling strategies span 6 categories (MLP dampening, contrast residual, late residual, attention dampening, MLP boosting, and full dampening) with coefficients $\alpha \in \{0.5, 0.7, 0.85, 0.9, 1.1\}$. ITI grid search covers $\lambda \in \{1.0, 1.5, 2.0, 2.5, 3.0\}$ and top\_k $\in \{3, 5, 10\}$.

\section{Methodology Details}
\label{app:methods}

\paragraph{Improved Causal Tracing (v2).} We extend \citet{meng2022locating} with adaptive noise calibration ($\sigma = \alpha \cdot \text{std}(\mathbf{E}(x))$), KL divergence metric ($\text{IE}(l) = (\text{KL}_{\text{corrupt}} - \text{KL}_{\text{patch}_l}) / \text{KL}_{\text{corrupt}}$, computed in float32), multi-run averaging ($n{=}5$ independent corruption runs), and head-level granularity via \texttt{hook\_z} patching.

\paragraph{Contrastive Activation Analysis.} Layer importance is computed as the normalized L2 distance: $\text{Imp}(l) = \| \bar{\mathbf{x}}_l^{\text{correct}} - \bar{\mathbf{x}}_l^{\text{halluc}} \|_2 / \max_l \| \bar{\mathbf{x}}_l^{\text{correct}} - \bar{\mathbf{x}}_l^{\text{halluc}} \|_2$, using last-token residual stream activations.

\section{Preliminary Results}
\label{app:preliminary}

\paragraph{Causal Tracing.} In Pythia-1.4B, MLP layer 0 is the top recovery layer for all 8 prompts (scores 0.66--0.99). Qwen2.5-0.5B shows a similar pattern with secondary peaks at layers 3--4. Attention tracing is distributed across early-to-middle layers with lower recovery scores (0.12--0.30).

\paragraph{Contrastive Analysis.} Top-5 contrastive layers: Qwen2.5-0.5B (22, 23, 21, 20, 19), Pythia-1.4B (23, 22, 21, 20, 19), Qwen2.5-7B (27, 26, 25, 24, 23). All models show late-layer concentration.

\paragraph{Head-Level Tracing.} In Pythia-1.4B, L6H12 (avg recovery 0.45) and L5H4 (0.45) are most critical. In Qwen2.5-0.5B, L2H4 (0.26) and L2H0 (0.23) show the highest scores. Critical heads are spread across early-to-middle layers, contrasting with late-layer contrastive importance.

Tables~\ref{tab:neg_scaling_05b}--\ref{tab:neg_iti} provide the full per-strategy intervention results summarized in Section~\ref{sec:negative_results}.

\begin{table}[htbp]
\centering
\small
\begin{tabular}{llccc}
\toprule
\textbf{Category} & \textbf{Strategy} & \textbf{Pre (\%)} & \textbf{Post (\%)} & \textbf{$\Delta$} \\
\midrule
\multirow{2}{*}{MLP Dampen} & $\alpha$=0.7 & 52.0 & 60.0 & $-$8.0 \\
 & $\alpha$=0.9 & 52.0 & 53.0 & $-$1.0 \\
\midrule
\multirow{2}{*}{Contrast Res.} & $\alpha$=0.9 & 52.0 & 50.0 & +2.0 \\
 & $\alpha$=0.7 & 52.0 & 67.0 & $-$15.0 \\
\midrule
Late Residual & $\alpha$=0.85 & 52.0 & 60.0 & $-$8.0 \\
\bottomrule
\end{tabular}
\caption{Representative activation scaling results on Chinese benchmark (Qwen2.5-0.5B, 100 samples). Only 2 of 20 strategies show positive effect ($\leq$2\%).}
\label{tab:neg_scaling_05b}
\end{table}

\begin{table}[htbp]
\centering
\small
\begin{tabular}{llccc}
\toprule
\textbf{Strategy} & \textbf{Config} & \textbf{Pre (\%)} & \textbf{Post (\%)} & \textbf{$\Delta$} \\
\midrule
Contrast Res. & $\alpha$=1.1 & 38.0 & 37.0 & +1.0 \\
Late Residual & $\alpha$=0.9 & 38.0 & 66.0 & $-$28.0 \\
\bottomrule
\end{tabular}
\caption{Activation scaling on Qwen2.5-7B. Best strategy achieves only +1\%; late residual suppression causes catastrophic degradation.}
\label{tab:neg_scaling_7b}
\end{table}

\begin{table}[H]
\centering
\small
\begin{tabular}{llcc}
\toprule
\textbf{Model} & \textbf{ITI Config} & \textbf{Truthful (\%)} & \textbf{$\Delta$} \\
\midrule
Pythia-1.4B & Baseline & 9.0 & --- \\
 & top3\_$\lambda$2.5 & 11.0 & +2.0 \\
Qwen2.5-0.5B & Baseline & 6.0 & --- \\
 & top3\_$\lambda$2.5 & 6.0 & 0.0 \\
\bottomrule
\end{tabular}
\caption{ITI grid search results (18 configurations). Best: Pythia +2\%, Qwen +0\%.}
\label{tab:neg_iti}
\end{table}

\begin{table}[H]
\centering
\small
\resizebox{\columnwidth}{!}{
\begin{tabular}{lccccc}
\toprule
\textbf{Method} & \textbf{top\_k} & \textbf{$\lambda$/coeff} & \textbf{MC1} & \textbf{MC2} & \textbf{$\Delta$MC1\%} \\
\midrule
Baseline & --- & --- & 0.2215 & 0.3921 & --- \\
\midrule
\multirow{9}{*}{\iti{}} & 3 & 1.0 & 0.2215 & 0.3921 & 0.0 \\
 & 3 & 2.0 & 0.2215 & 0.3921 & 0.0 \\
 & 3 & 3.0 & 0.2215 & 0.3921 & 0.0 \\
 & 5 & 1.0 & 0.2215 & 0.3924 & 0.0 \\
 & 5 & 2.0 & 0.2203 & 0.3926 & $-$0.5 \\
 & 5 & 3.0 & 0.2215 & 0.3929 & 0.0 \\
 & 10 & 1.0 & 0.2326 & 0.4005 & +5.0 \\
 & 10 & 2.0 & 0.2362 & 0.4092 & +6.6 \\
 & 10 & 3.0 & \textbf{0.2436} & \textbf{0.4178} & \textbf{+10.0} \\
\midrule
\multirow{3}{*}{\dola{}} & \multicolumn{2}{c}{early (static)} & 0.2326 & 0.4371 & +5.0 \\
 & \multicolumn{2}{c}{mid (static)} & 0.2448 & 0.4555 & +10.5 \\
 & \multicolumn{2}{c}{dynamic} & \textbf{0.2778} & \textbf{0.4822} & \textbf{+25.4} \\
\midrule
\multirow{18}{*}{\caa{}} & 3 & 0.5 & 0.2215 & 0.3921 & 0.0 \\
 & 3 & 1.0 & 0.2215 & 0.3921 & 0.0 \\
 & 3 & 1.5 & 0.2215 & 0.3921 & 0.0 \\
 & 3 & 2.0 & 0.2215 & 0.3921 & 0.0 \\
 & 3 & 3.0 & 0.2215 & 0.3921 & 0.0 \\
 & 3 & 5.0 & 0.2215 & 0.3920 & 0.0 \\
 & 5 & 0.5 & 0.2215 & 0.3922 & 0.0 \\
 & 5 & 1.0 & 0.2215 & 0.3924 & 0.0 \\
 & 5 & 1.5 & 0.2215 & 0.3925 & 0.0 \\
 & 5 & 2.0 & 0.2203 & 0.3926 & $-$0.5 \\
 & 5 & 3.0 & 0.2215 & 0.3929 & 0.0 \\
 & 5 & 5.0 & 0.2215 & 0.3934 & 0.0 \\
 & 10 & 0.5 & 0.2252 & 0.3963 & +1.7 \\
 & 10 & 1.0 & 0.2326 & 0.4005 & +5.0 \\
 & 10 & 1.5 & 0.2362 & 0.4047 & +6.6 \\
 & 10 & 2.0 & 0.2362 & 0.4092 & +6.6 \\
 & 10 & 3.0 & 0.2436 & 0.4178 & +10.0 \\
 & 10 & 5.0 & \textbf{0.2558} & \textbf{0.4338} & \textbf{+15.5} \\
\bottomrule
\end{tabular}
}
\caption{Complete intervention results on TruthfulQA (Qwen2.5-7B, 817 samples). All 30 configurations from the main text. Key finding: \iti{} and \caa{} show \emph{zero} improvement at top\_k$\in\{3,5\}$ but substantial gains at top\_k=10, supporting the layer-count threshold hypothesis.}
\label{tab:full_intervention_results}
\end{table}

\section{CrystalBoost: Mechanistic Validation}
\label{app:crystalboost}

To test whether mechanistic insight into crystallization can directly inform intervention design, we construct \crystalboost{}, a dual-stage intervention: early layers (0--$\lfloor 0.5n \rfloor$) suppress surface pattern directions, late layers ($\lfloor 0.8n \rfloor$--$n$) amplify factual crystallization directions, with Gaussian-weighted coefficients at the boundary ($\sim$90\% depth).

\paragraph{Why \crystalboost{} underperforms \dola{} on high-crystallization models.} The +2.2\% gain on Qwen vs.\ \dola{}'s +26.4\% does not indicate failure of FEP theory---rather, it reveals a fundamental \emph{space mismatch}. \dola{} operates in \textbf{logit space} where crystallization is directly visible: post-crystallization logits contain the correct answer, enabling effective contrast. \crystalboost{} steers in \textbf{activation space} where 85.9\% of correct answers remain implicit (never entering top-10 at any intermediate layer). On Llama (71.0\% crystallization), where more knowledge is activation-visible before the final layer, \crystalboost{} outperforms \dola{} (+8.4\% vs.\ +1.9\%). This pattern \emph{validates} FEP theory: activation-space methods are most effective when crystallization is moderate, while logit-space methods become preferable when crystallization is extreme.

\begin{table}[htbp]
\centering
\small
\resizebox{\columnwidth}{!}{
\begin{tabular}{llccc}
\toprule
\textbf{Model} & \textbf{Method} & \textbf{MC1} & \textbf{$\Delta$\%} & \textbf{vs.\ DoLa} \\
\midrule
\multirow{3}{*}{Qwen2.5-7B} & Baseline & 0.2179 & --- & --- \\
 & \dola{} dynamic & 0.2754 & +26.4 & --- \\
 & \crystalboost{} (best) & 0.2228 & +2.2 & loses \\
\midrule
\multirow{3}{*}{Llama-3.1-8B} & Baseline & 0.1897 & --- & --- \\
 & \dola{} dynamic & 0.1934 & +1.9 & --- \\
 & \crystalboost{} (best) & \textbf{0.2056} & \textbf{+8.4} & \textbf{wins} \\
\bottomrule
\end{tabular}
}
\caption{\crystalboost{} evaluation (grid search, 5 configurations). Consistent positive improvements (+2.2\% Qwen, +8.4\% Llama) validate the predictive power of FEP theory. Minor baseline variance from Table~\ref{tab:mc_results} reflects different evaluation batches.}
\label{tab:crystalboost}
\end{table}

\section{FEP Distribution Details}
\label{app:fep_dist}

Table~\ref{tab:fep_distribution} and Figure~\ref{fig:fep_dist} in the main text present the full FEP distribution data across architectures. Key observations: (1) Qwen2.5-7B exhibits the most concentrated distribution, with 85.9\% of samples showing FEP at the final layer (layer 27--28 of 28); (2) Llama-3.1-8B shows a broader distribution centered at layers 25--28, reflecting its lower crystallization rate (70.4\%); (3) Mistral-7B displays the widest spread (FEP range: layers 15--28), consistent with sliding window attention enabling earlier information routing; (4) Pythia-6.9B and Gemma-7B both show high crystallization rates (93.4\% and 89.8\%) with narrow FEP distributions concentrated at their respective final layers. Across all architectures, the FEP distribution is unimodal and right-skewed, confirming that Late Crystallization is a general phenomenon rather than an artifact of any single architecture.

\begin{figure*}[!b]
    \centering
    \includegraphics[width=\textwidth]{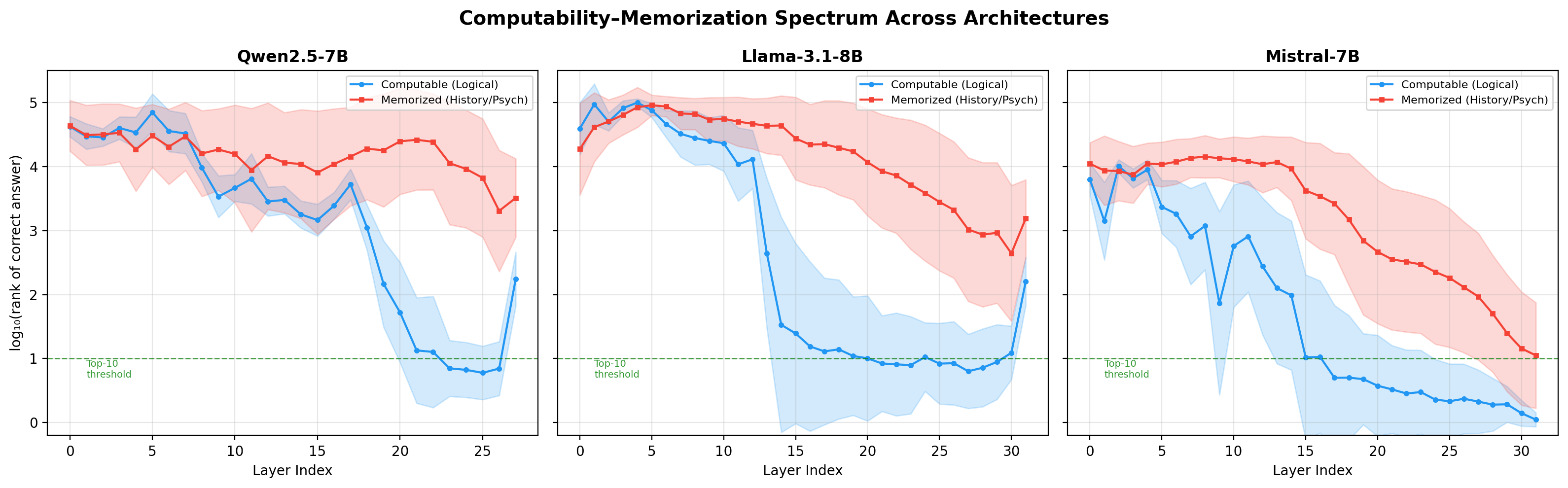}
    \caption{Computability--Memorization Spectrum across architectures. Blue: computable knowledge (logical reasoning); Red: memorized knowledge (history, psychology). Computable categories cross the top-10 rank threshold at earlier layers across all three models.}
    \label{fig:spectrum}
\end{figure*}

\section{Top-$k$ Sensitivity Analysis}
\label{app:topk}

\begin{table}[htbp]
\centering\small
\begin{tabular}{rcccc}
\toprule
\textbf{$k$} & \textbf{Mean FEP} & \textbf{$\sigma$} & \textbf{Late Crystal} & \textbf{FEP Depth} \\
\midrule
1 & 28.0 & 0.0 & 100.0\% & 100.0\% \\
3 & 27.9 & 0.4 & 98.3\% & 99.8\% \\
5 & 27.6 & 1.4 & 92.9\% & 98.7\% \\
10 & 27.3 & 1.8 & 85.9\% & 97.5\% \\
20 & 27.0 & 2.1 & 80.2\% & 96.5\% \\
50 & 26.6 & 2.6 & 72.7\% & 95.0\% \\
100 & 26.0 & 3.4 & 64.7\% & 93.0\% \\
\bottomrule
\end{tabular}
\caption{FEP sensitivity to $k$ (Qwen2.5-7B, 817 samples). Late Crystallization is robust: even at $k{=}100$, 64.7\% of answers never appear at intermediate layers. FEP Depth stays $>$93\% across all thresholds.}
\label{tab:topk_sensitivity}
\end{table}

\section{Cross-Benchmark FEP Validation (MMLU)}
\label{app:mmlu}

\begin{table}[htbp]
\centering\small
\begin{tabular}{lccc}
\toprule
\textbf{Benchmark} & \textbf{$n$} & \textbf{Mean FEP} & \textbf{Late Crystal} \\
\midrule
TruthfulQA & 817 & 27.3$\pm$1.8 & 85.9\% \\
MMLU (all) & 1{,}200 & 27.9$\pm$0.5 & 98.2\% \\
\midrule
\quad STEM & 450 & 28.0$\pm$0.4 & 99.1\% \\
\quad Humanities & 250 & 28.0$\pm$0.4 & 98.0\% \\
\quad Social Sciences & 250 & 27.9$\pm$0.7 & 97.2\% \\
\quad Other & 250 & 27.9$\pm$0.6 & 98.0\% \\
\bottomrule
\end{tabular}
\caption{Cross-benchmark FEP validation (Qwen2.5-7B). Late Crystallization is even more pronounced on MMLU (98.2\%) than TruthfulQA (85.9\%), and consistent across all MMLU subject groups (97.2--99.1\%).}
\label{tab:mmlu_fep}
\end{table}

\section{Computability--Memorization Spectrum Details}
\label{app:spectrum}

Figure~\ref{fig:spectrum} visualizes the Computability--Memorization Spectrum across three architectures. Categories span a wide FEP range: Logical Falsehood crystallizes earliest (mean FEP=22.1, $\sigma$=2.6), indicating that logically computable knowledge crosses the top-10 rank threshold at intermediate layers. In contrast, purely memorized categories---History, Psychology, and Weather---show FEP=28.0 with \emph{zero} variance, meaning every sample's correct answer remains invisible until the final layer. This gradient is consistent across Qwen2.5-7B, Llama-3.1-8B, and Mistral-7B, with computable categories systematically shifted leftward (earlier crystallization) relative to memorized ones. The pattern suggests that category-adaptive interventions---applying stronger steering to high-FEP memorized categories while preserving the natural emergence of computable knowledge---could yield further improvements. Full per-category FEP statistics are available upon request.

\end{document}